\PassOptionsToPackage{colorlinks=true,allcolors=black}{hyperref}

\documentclass[11pt]{article}

\usepackage[T1]{fontenc}
\usepackage{lmodern}

\usepackage{setspace}
\usepackage{amsmath, amssymb}
\usepackage{array}
\usepackage[table]{xcolor}
\usepackage{colortbl}
\usepackage{booktabs}
\usepackage{graphicx}

\usepackage{amsthm}

\theoremstyle{plain}
\newtheorem{theorem}{Theorem}[section]

\newtheorem{constraint}[theorem]{Constraint}

\newtheorem{assumption}[theorem]{Assumption}

\theoremstyle{definition}
\newtheorem{definition}[theorem]{Definition}
\newtheorem{example}[theorem]{Example}

\theoremstyle{remark}
\newtheorem{remark}[theorem]{Remark}
\newtheorem{note}[theorem]{Note}

\newtheorem*{constraint*}{Constraint}
\newtheorem*{definition*}{Definition}
\newtheorem*{example*}{Example}
\newtheorem*{remark*}{Remark}
\newtheorem*{note*}{Note}

\usepackage{url}
\usepackage{natbib}
\bibpunct{(}{)}{;}{a}{}{ }

\usepackage{hyperref}
\usepackage{bookmark}

\usepackage{authblk}

\usepackage{microtype}

\usepackage[most]{tcolorbox}

\usepackage{enumitem}
\setlist[itemize]{itemsep=0.2em, topsep=0.2em, parsep=0em, partopsep=0em}
\setlist[enumerate]{itemsep=0.2em, topsep=0.2em, parsep=0em, partopsep=0em}

\providecommand{\keywords}[1]{\vspace{0.5em}\noindent\textbf{Keywords: }\small #1}

\newcommand{\DefRef}[2]{\hyperref[#1]{Definition (#2)}}
\newcommand{\AssumpRef}[2]{\hyperref[#1]{Assumption (#2)}}
\newcommand{\ConstRef}[2]{\hyperref[#1]{Constraint (#2)}}
\newcommand{\ExRef}[2]{\hyperref[#1]{Example (#2)}}
\newcommand{\RemRef}[2]{\hyperref[#1]{Remark (#2)}}
\newcommand{\NoteRef}[2]{\hyperref[#1]{Note (#2)}}
\newcommand{\SecRef}[1]{Section~\ref{#1}}

\newcommand{\Substrate}{\mathcal{S}}
\newcommand{\SubstrateRef}{\mathcal{S}_{\mathrm{ref}}}
\newcommand{\Frameworks}{\mathbb{F}}
\newcommand{\Framework}{F}
\newcommand{\FrameworkOne}{F_1}
\newcommand{\FrameworkTwo}{F_2}
\newcommand{\entails}{\vdash}
\newcommand{\notentails}{\nvdash}
\newcommand{\Asserts}{\mathsf{Asserts}}
\newcommand{\pred}[1]{\mathit{#1}}

\title{
Neutral Substrates:\\
A Design Constraint for Shared Records\\
Under Persistent Interpretive Disagreement
}

\author{Denise M. Case}
\affil{\small Department of Computer Science and Information Systems\\
Northwest Missouri State University, Maryville, MO, USA\\}
\date{}

\begin{document}
\maketitle
\vspace{-2em}

\begin{abstract}
  Shared accountability records must often remain useful to parties
  who disagree about causation, responsibility, and normative interpretation.
  Suppressing these claims would erase information needed for accountability;
  asserting one disputed interpretation as foundational would privilege it.
  A neutral substrate instead separates shared referential commitments
  from attributed interpretive assertions.

  This paper formalizes that separation through admissible frameworks,
  referential commitments, permitted attributions, interpretive
  non-commitment, and extension stability.
  Under Referential Common Ground,
  restricting every substrate-layer commitment to a referential commitment
  or a permitted attribution proposition is sufficient for semantic neutrality
  relative to every admissible framework.
  It is not a design-time guarantee: such a guarantee requires a permitted basis
  establishing the neutrality result, rather than only its truth.

  The essential assumption is joint compatibility of the complete
  referential commitment set with each admissible framework.
  Separate compatibility of individual commitments is not sufficient,
  and admissible frameworks need not be mutually consistent.
  Furthermore, the formal referential interfaces do not independently
  establish the identity criteria, denotation relations, attributional bases,
  or interpretive adequacy required by a particular application.
  When those premises cannot be justified, neutrality cannot be certified
  by the present construction.

  The contribution is a precise conditional layer restriction and
  an explicit account of its semantic and scientific boundaries.
  Attribution and provenance are established representational mechanisms;
  their presence alone cannot establish referential fixing.
\end{abstract}


\begin{note*}[Citation Convention]
  Definitions, assumptions, constraints, and examples in this paper carry stable
  names (for example, \texttt{se100.def.Substrate}).
  This paper cites statements by stable name rather than by number,
  so citations remain valid even if numbering changes.
\end{note*}


\bigskip
\noindent\textbf{Preview of Main Constraint.}
\emph{
  Assume that the complete set of referential commitments is
  consistent when combined with each admissible framework.
  If all substrate-layer commitments belong to that set or are
  permitted attribution propositions fixed by it, then the
  substrate satisfies extension stability and interpretive
  non-commitment.
  A separate permitted design-time guarantee basis is needed
  to establish neutrality by design.
}

\medskip
\noindent
\emph{Informally.} A shared record can preserve what sources asserted
without adopting the asserted content.
The claim is conditional on soundness and compatibility
of the adopted referential basis;
it does not establish those premises for an actual system.

\section{Introduction}
\label{sec:introduction}

Accountability records are rarely interpreted by only one party.
A record may be used by institutions, auditors, affected individuals,
regulators, courts, researchers, vendors, advocates, or future maintainers.
These parties may share enough reference to identify
the person, event, decision, instrument, place, file, policy, rule,
or institutional artifact at issue.
At the same time, they may disagree, possibly permanently,
about what caused what, who was responsible,
whether a decision was justified, which rule applied,
or which normative standard should control.

This creates a structural design problem.
A shared record must remain usable by multiple interpretive frameworks,
including frameworks that permanently disagree.
If the shared layer commits to an interpretation that is incompatible
with some admissible framework, extension stability fails.
Even where no conflict has been exhibited, an unqualified contested
commitment cannot be certified by the permitted-layer restriction
without establishing the corresponding compatibility premises.

The purpose of this paper is to define the substrate-layer constraint
sufficient to avoid that failure.

The central claim is:

\begin{quote}
  In the permitted-layer construction, contested causal or normative
  claims can be preserved as attributed, provenance-bearing assertions
  rather than adopted as unqualified substrate commitments.
  The neutrality result depends on justified referential fixing
  and compatibility with each admissible framework.
\end{quote}

The mechanism of neutrality is the distinction between
asserting a proposition and representing that some identified source $x$
asserted it.
This mechanism avoids two failures.
It does not suppress contested content,
because the attributed claim remains in the record.
It does not adopt contested content,
because the substrate commits to the attribution proposition
$\Asserts(x,\varphi)$,
not to the asserted proposition $\varphi$.

The mechanism is not new (\SecRef{sec:related-work}).
The contribution is a conditional sufficiency result for semantic neutrality
together with the distinction between that result and a design-time guarantee.
The permitted classes specify one justified route to neutrality
under an independently warranted common-ground assumption;
they do not establish that every possible neutral design has this form.

The paper proceeds as follows.
\SecRef{sec:related-work} credits the foundations and states the contribution.
\SecRef{sec:problem-setting} states the problem.
\SecRef{sec:definitions} gives the core definitions.
\SecRef{sec:neutrality} states and justifies the neutrality constraint.
\SecRef{sec:mechanism} explains the reification mechanism.
\SecRef{sec:example} gives a worked example.
\SecRef{sec:boundary} states the boundary condition under which the construction does not certify neutrality.
\SecRef{sec:non-goals} states the scope and limits of the constraint.
\SecRef{sec:discussion} discusses the practical significance and limits of the constraint.
\SecRef{sec:conclusion} concludes.

\section{Related Work and Contribution}
\label{sec:related-work}

The representational mechanism this paper relies on is not novel.
Prior work developed machinery for
recording that a source asserts a proposition
rather than asserting the proposition itself.
This is standard in knowledge representation,
logical AI,
argumentation theory,
and applied ontology.

In knowledge representation on the web,
RDF reification \citep{hayes2004rdf}
and named graphs \citep{carroll2005named}
provide standard machinery often used for
statements about statements.
The W3C provenance ontology
\citep{lebo2013prov} standardizes attribution of content to
agents, activities, and sources.
The form $\Asserts(x,\varphi)$ used in this paper,
with $x$ identifying the source to which the assertion is attributed,
follows that representational pattern;
nothing representational is added.

In logical AI, \citet{mccarthy1993} treats truth as relative to a context via
$\mathit{ist}(c,p)$.
Framework-relative derivability
($\Substrate \cup \Framework \entails p$)
uses the same general idea in a narrower way.
Purpose is the difference:
context logic supports reasoning across contexts,
whereas the substrate defined here deliberately abstains from
cross-framework commitments at its foundational layer.
Assumption-based truth maintenance \citep{dekleer1986atms}
likewise maintains conclusions relative to assumption sets
and keeps mutually inconsistent worlds simultaneously available.
Extension stability, defined below,
is a record-level analogue of that discipline:
an ATMS is a reasoning architecture,
whereas this constraint is a design condition on records.

In argumentation theory, \citet{dung1995} showed how to compute acceptability
among conflicting arguments.
The aim here is complementary and prior:
argumentation frameworks adjudicate among conflicting claims,
whereas a neutral substrate refuses adjudication at the base layer,
preserving the conflicting claims with attribution.
A neutral substrate is the kind of shared input over
which a Dung-style evaluation could later be run by a framework,
not by the substrate.

Standpoint logic \citep{gomezalvarez2021standpoint} is the closest recent
knowledge-representation relative.
It extends a base language with labelled modalities,
$\Box_s\varphi$ (``according to $s$, it is unequivocal that $\varphi$''),
so that knowledge bases holding conflicting perspectives can
be reasoned over jointly rather than merged.
The two address different questions.
Standpoint logic is a representation and reasoning language
over a given set of standpoints;
the constraint here concerns which commitments
a shared foundational layer may make so that it
remains consistent with every admissible framework,
including frameworks not known at design time
(\NoteRef{se100.note.FrameworkClass}{The Framework Class}).
The attribution form $\Asserts(x,\varphi)$ records an attributed assertion
with provenance rather than a perspective-relative truth condition.
The approaches are possibly complementary:
a standpoint-logic reasoner could serve as the framework-side language,
with the layer restriction as a discipline on the shared base.

Two older distinctions underwrite the whole construction.
Speech act theory distinguishes asserting a proposition
from reporting that someone asserted it \citep{searle1969}.
The analysis of ontological commitment \citep{guarino1998}
makes precise the idea that a representation commits its
users to a view of the domain.
The substrate-layer commitment defined below is
a record-level application of that idea.

Within applied ontology,
the descriptions-and-situations pattern of DOLCE+DnS Ultralite
represents descriptions, situations, and interpretive stances
as first-class modeling elements distinct
from the states of affairs they describe
\citep{gangemi2003descriptions}.
The present constraint is compatible with that pattern and
can be read as a condition on which layer a commitment occupies:
reference and permitted attribution at the substrate,
causal and normative interpretation in framework-dependent descriptions.
This paper does not introduce a new reification mechanism.
Its contribution is a design-time constraint on the shared foundational layer:
when neutrality under persistent interpretive disagreement is required,
that layer is restricted
to referential commitments and permitted attribution propositions.

This paper is also adjacent to Floridi's method of levels of abstraction
\citep{floridi2008loa}.
The analysis here is stated as a representational-layer constraint
rather than as a level-of-abstraction claim:
the issue is not only which abstraction an interpreter selects,
but which commitments the shared record adopts at its foundational layer.

\textbf{The contribution.}
Prior literatures provide the ingredients but do not state
an explicit condition on the foundational layer of a shared accountability record.
This paper assembles those ingredients into a conditional
foundational-layer restriction and states the assumptions needed for its
semantic sufficiency.
The sufficiency argument uses only the structural properties
of consequence stated in \NoteRef{se100.note.ConsequenceSystem}{Consequence System},
so it does not depend on a particular logic.

Reification and provenance provide the mechanism.
Context logic and truth maintenance provide the relativized-derivability reading.
Argumentation theory provides the downstream consumer.
Speech act theory and ontological commitment provide the underlying distinctions.

The constraint combines an admissibility gate for frameworks,
a contestability assumption for causal and normative content,
a common-ground assumption for reference,
and compatibility between the complete substrate commitment set and
each admissible framework.
It also states the admissibility and grounding conditions that make it usable,
including the boundary at which it fails.

Each ingredient is known.
The addition is the constraint and the discipline of stating
the conditions under which it does not apply.

\section{Problem Setting}
\label{sec:problem-setting}

A shared record can fail neutrality in two opposed ways.

\begin{enumerate}
  \item It can omit accountability-relevant claims entirely.
        This avoids adopting those claims as object-level substrate-layer commitments,
        but it also suppresses the information needed to understand
        who claimed what, under what authority, and on what basis.
        A record that suppresses contested claims cannot support accountability
        for those claims.

  \item It can make a substrate-layer commitment
        to a contested causal or normative proposition.
        This preserves the content of the claim,
        but strips away the attribution that makes the claim contestable.
        Such a record may appear complete, but is structurally non-neutral.
\end{enumerate}

A neutral substrate must avoid both failures.
It must preserve contested claims without adopting them
as object-level substrate-layer commitments.

This paper uses a role-based taxonomy of \emph{substrate-carried content}:

\begin{enumerate}
  \item \textbf{Referential content}: entities, occurrences, institutional
        artifacts, identifiers, timestamps, provenance, and the relations needed
        to fix what is being referred to.
  \item \textbf{Attribution content}: propositions of the form
        $\Asserts(x,\varphi)$, meaning that some
        framework, source, agent, institution, record, or document $x$
        asserts a claim $\varphi$ about the referents.
  \item \textbf{Object-level interpretive content}: the asserted propositions
        $\varphi$ themselves about the referents, rather than attribution
        propositions about those assertions.
        In this paper, the contested content of concern
        is object-level causal or normative propositions;
        the formal restriction in \SecRef{sec:neutrality}
        applies to all substrate-layer commitments.
\end{enumerate}

This taxonomy distinguishes representational roles,
not mutually exclusive kinds of propositions.
When the substrate commits to content of any of these kinds,
that commitment is a substrate-layer commitment in the sense of
\DefRef{se100.def.SubstrateCommitment}{Substrate-Layer Commitment}.
The substrate may entail a referential proposition,
an attribution proposition,
or an object-level interpretive proposition.
The three kinds differ in \emph{what} is committed,
not in \emph{whether} the substrate commits.

The permitted-layer construction accepts referential commitments and
attributions whose bases are fixed by those commitments.
A contested causal or normative assertion can be preserved without adopting
its content by recording a permitted attribution rather than committing
to the object-level proposition itself.

The restriction's supply of semantic neutrality depends on whether the referential
commitment set is jointly compatible with every admissible framework.
It does not itself identify which records genuinely fix references
or supply a design-time warrant for their compatibility.

\section{Definitions}
\label{sec:definitions}

\subsection*{Substrates and Commitments}

\begin{definition}[Substrate]
  \label{se100.def.Substrate}
  A \emph{substrate} $\Substrate$ is a shared representational base providing
  stable reference for entities, occurrences, and institutional artifacts
  across a class of admissible interpretive frameworks $\Frameworks$.
  Stable reference consists of individuation, co-reference, and persistence.
\end{definition}

A substrate is not the whole record system.
It is the foundational layer on which further interpretive claims may be layered,
including causal or normative interpretations,
evidentiary assessments, explanatory accounts, and institutional determinations.
Its purpose is to make shared reference possible without
forcing agreement about every interpretation of what is referenced.
A substrate may identify that there is an event, entity, decision,
document, instrument, rule, policy, observation, or institutional artifact, and
it may preserve identifiers, timestamps, provenance, and persistence relations
among them.
Whether some event caused another,
whether some decision was justified, or
whether some rule ought to apply
are not commitments of the substrate
just because the substrate identifies
the relevant event, decision, rule,
or institutional artifact.


\begin{definition}[Substrate-Layer Commitment]
  \label{se100.def.SubstrateCommitment}
  A substrate $\Substrate$ \emph{commits} to a proposition $p$ when
  $\Substrate \entails p$;
  that is, $p$ is asserted by the substrate independently of any interpretive
  framework.
\end{definition}

\begin{note}[Consequence System]
  \label{se100.note.ConsequenceSystem}
  The arguments below use only three properties of $\entails$:
  (i)~every member of a theory is entailed by it;
  (ii)~generalized cut: if $T$ entails every member of $U$ and $U$
  entails $p$, then $T$ entails $p$ (monotonicity follows); and
  (iii)~if a theory entails both $p$ and $\neg p$, it entails $\bot$.
  No explosion, excluded middle, negation introduction, double negation,
  completeness, decidability, or compactness is assumed.
\end{note}

\subsection*{Proposition Roles}

\begin{note}[Causal and Normative Content as Primitive Classifications]
  \label{se100.note.CausalNormative}
  This paper treats \emph{causal proposition} and \emph{normative proposition}
  as primitive content classifications,
  not as terms reduced here to a formal decision procedure.
  As a working guide,
  a proposition is causal if it asserts that one event,
  state, action, condition, or process brought about, produced, prevented, or
  contributed to another.
  A proposition is normative if it asserts that something was justified,
  permitted, required, prohibited, correct, incorrect, compliant, or in
  violation of a rule, standard, policy, or norm.

  The classification is not by surface vocabulary.
  An innocuous-looking field can encode a causal or normative commitment,
  and causal or normative words can appear inside an attributed assertion
  without the substrate committing to the asserted causal or normative
  proposition.
  The constraint applies once the relevant content classification is fixed
  for the accountability context.
\end{note}


\begin{definition}[Attribution Proposition]
  \label{se100.def.AttributionProposition}
  An \emph{attribution proposition} is a proposition of the form
  $\Asserts(x,\varphi)$,
  meaning that some framework, source, agent, institution, record, or document $x$
  asserts proposition $\varphi$.

  A substrate-layer commitment to $\Asserts(x,\varphi)$ is a
  commitment to the attribution,
  not to the asserted proposition $\varphi$.
\end{definition}


\begin{definition}[Object-Level Interpretive Proposition]
  \label{se100.def.ObjectLevelInterpretiveProposition}
  An \emph{object-level interpretive proposition} is an asserted proposition
  $\varphi$ itself,
  about the referents fixed by the substrate,
  rather than an attribution proposition of the form
  $\Asserts(x,\varphi)$.

  An \emph{object-level causal or normative proposition} is
  an object-level interpretive proposition whose content is causal or normative
  (\NoteRef{se100.note.CausalNormative}{Causal and Normative Content as
    Primitive Classifications}).
\end{definition}

The core distinction is that the substrate may commit to
$\Asserts(x,\varphi)$
without committing to $\varphi$.
Neutrality does not require suppressing causal or normative information.
It requires keeping causal and normative information at the level
of attribution propositions rather than substrate-layer commitments
to the asserted propositions.

\begin{definition}[Object-Level Causal or Normative Commitment]
  \label{se100.def.ObjectLevelCausalNormativeCommitment}
  An \emph{object-level causal or normative commitment} is a substrate-layer
  commitment to an object-level causal or normative proposition,
  rather than to an attribution proposition about that proposition.
\end{definition}

\subsection*{Reference}

\begin{definition}[Referential Regime]
  \label{se100.def.ReferentialRegime}
  A \emph{referential regime} is the triple of
  \begin{enumerate}
    \item individuation conditions,
    \item co-reference conditions, and
    \item persistence conditions
  \end{enumerate}
  by which a substrate fixes and tracks entities, occurrences, and institutional
  artifacts.
\end{definition}

Individuation specifies conditions for treating a presentation as one
unit and for judging sameness across presentations.
Co-reference concerns references denoting the same referent.
Persistence concerns strict identity across settings and must be
distinguished from succession, derivation, and counterpart relations.
These questions are related but cannot be answered by stipulating
a carrier identifier or an equivalence relation.

The referential regime here is an abstract interface for these accounts.
It does not construct an independently grounded criterion for any
intended referent kind, or establish their mutual compatibility.
That substantive work is required before their conclusions can be
adopted as fixed referential commitments.

\begin{definition}[Referential Commitments]
  \label{se100.def.ReferentialCommitments}
  Let $\SubstrateRef$ denote the \emph{referential commitments} of $\Substrate$.
  These are the substrate-layer commitments fixed by its referential regime,
  including identifiers;
  the typing of entities, occurrences, and institutional artifacts;
  timestamps;
  provenance;
  and referential relations among them.
\end{definition}

Identifiers, timestamps, typing declarations, provenance records, and
relations are possible materials for establishing referential commitments;
their presence alone is not evidence that a proposition is referentially fixed.
A concrete realization must justify the connection between those records,
the referential regime, and the object-language propositions claimed to be fixed.

\subsection*{Frameworks}

\begin{definition}[Admissible Framework]
  \label{se100.def.AdmissibleFramework}
  A framework $\Framework$ is \emph{admissible} if it satisfies three conditions:
  \begin{enumerate}
    \item \textbf{Internal consistency}: $\Framework \notentails \bot$.
    \item \textbf{Evidentiary grounding}: where $\Framework$ makes empirical,
          causal, normative, or other interpretive claims, it identifies the evidence,
          source, method, measurement, observation, record, rule, standard, or
          document on which those claims rely.
    \item \textbf{Documented interpretive function}: $\Framework$ is presented as
          an interpretive function with a
          named source, documented scope, and citable basis.
  \end{enumerate}
\end{definition}

Admissibility is not agreement.
Two admissible frameworks may disagree about
causation, responsibility, justification, applicability, or normative significance.
Admissibility does not determine which framework should prevail.
It specifies the minimum conditions under which a framework can enter
the shared record as an identifiable, scoped, and citable interpretive position.

The admissibility gate is not neutral, and this paper does not claim that.
Requiring evidentiary grounding and a documented interpretive function
(Conditions~2 and~3) embeds design choices about what counts as
a framework that can enter the shared record.

These conditions do not adjudicate the framework's claims.
They make the framework available for representation in the record.
Any further claim about whether a framework's basis is sufficient,
whether its interpretation should govern,
or whether its use is appropriate in a particular dispute
must be represented as an attributed claim with provenance,
not as a substrate-layer commitment.

\begin{note}[The Framework Class $\Frameworks$]
  \label{se100.note.FrameworkClass}
  Throughout, $\Frameworks$ denotes the
  class of \emph{all} admissible interpretive frameworks.
  Admissibility is an intrinsic property
  (\DefRef{se100.def.AdmissibleFramework}{Admissible Framework}),
  so membership in $\Frameworks$ does not depend on designer knowledge.
  The class is open and not enumerable at design time:
  new admissible interpretive frameworks can be authored at any
  point in a record's lifetime.
  ``Extension of $\Frameworks$'' below refers to
  such frameworks becoming known and entering consideration,
  not to change in the intrinsic class.
  This sense of extension is distinct from the
  substrate-with-framework extension $\Substrate \cup \Framework$ named in
  \DefRef{se100.def.ExtensionStability}{Extension Stability}.
\end{note}

\begin{definition}[Permitted Attribution Proposition]
  \label{se100.def.PermittedAttributionProposition}
  An attribution proposition $\Asserts(x,\varphi)$ is
  \emph{permitted at the foundational layer} if the attributional basis for
  $x$'s assertion of $\varphi$ is fixed by the substrate's referential
  commitments $\SubstrateRef$.

  The attributional basis includes the source, assertion occurrence,
  provenance, and content reference needed to identify what was asserted,
  by whom, and under what record basis.

  Here, fixed by $\SubstrateRef$ means determined by it:
  a permitted attribution proposition $\Asserts(x,\varphi)$ is among the
  substrate-layer commitments determined by $\SubstrateRef$; that is,
  \[
    \SubstrateRef \entails \Asserts(x,\varphi) .
  \]
  Committing to it commits the substrate to the attribution by $x$,
  not to the asserted proposition $\varphi$.
\end{definition}

\begin{definition}[Framework-Variant Proposition]
  \label{se100.def.FrameworkVariant}
  A proposition $p$ is \emph{framework-variant}
  with respect to substrate $\Substrate$ and framework class $\Frameworks$
  if there exist admissible frameworks $\FrameworkOne, \FrameworkTwo \in \Frameworks$
  such that
  \[
    \Substrate \cup \FrameworkOne \entails p
    \quad\text{and}\quad
    \Substrate \cup \FrameworkTwo \entails \neg p
  \]
\end{definition}

Framework variance captures disagreement that arises
not from different referential bases,
but from different admissible interpretations
layered on top of the same referential base.
A proposition accepted under one framework
and rejected under another is
not suitable as an unqualified substrate-layer commitment.

\begin{definition}[Framework-Invariant Proposition]
  \label{se100.def.FrameworkInvariant}
  A proposition $p$ is \emph{framework-invariant} with respect to substrate
  $\Substrate$ and framework class $\Frameworks$
  if, for every admissible framework $\Framework \in \Frameworks$,
  \[
    \Substrate \cup \Framework \cup \{p\} \notentails \bot .
  \]
  Equivalently,
  $p$ can be added to the substrate and
  remain compatible with every admissible framework.
\end{definition}

Framework variance
(\DefRef{se100.def.FrameworkVariant}{Framework-Variant Proposition})
and framework invariance are stated over the same shared base
$\Substrate \cup \Framework$.
They are opposed in one direction:
if $p$ is framework-variant with respect to $\Substrate$ and $\Frameworks$,
then some admissible $\Framework$ gives
$\Substrate \cup \Framework \entails \neg p$.
Adding $p$ to that same base yields
$\Substrate \cup \Framework \cup \{p\} \entails \bot$,
so $p$ is not framework-invariant.

\begin{remark}[Framework Invariance and Refutation]
  \label{se100.remark.FrameworkInvariantRefutation}
  Say that an admissible framework $\Framework$ \emph{refutes} $p$
  if $\Substrate \cup \Framework \entails \neg p$.
  Framework invariance implies that no admissible framework refutes $p$:
  if $\Substrate \cup \Framework \entails \neg p$,
  then $\Substrate \cup \Framework \cup \{p\}$ entails both $p$ and $\neg p$,
  hence $\bot$.

  The converse requires negation introduction:
  if $T \cup \{q\} \entails \bot$, then $T \entails \neg q$.
  This paper does not assume it.
  Framework invariance is therefore defined by consistency,
  not by non-refutation.
\end{remark}

\begin{definition}[Framework-Compatible Commitment Set]
  \label{se100.def.FrameworkCompatibleCommitmentSet}
  A set of substrate-layer commitments $C$ is \emph{framework-compatible}
  with respect to $\Frameworks$ if, for every admissible framework
  $\Framework \in \Frameworks$,
  \[
    C \cup \Framework \notentails \bot .
  \]
\end{definition}

Compatibility is evaluated for the commitment set \emph{jointly},
not separately for each member.
In a classical instance,
a proposition and its negation may each be individually consistent
with an otherwise empty commitment set, while their conjunction is not.
Consequently, pointwise screening of proposed referential commitments
cannot establish framework compatibility of their union.

Compatibility with each admissible framework also does not require
all admissible frameworks to be mutually compatible.
Finally, openness of the framework class does not imply
that every identity proposition not entailed by the substrate
is refuted by some admissible framework.
No such criterion-openness assumption is adopted here.

Framework invariance is a positive guarantee,
rather than absence of observed disagreement.
A proposition may fail to be framework-invariant
even when it is not framework-variant.
For example,
suppose no admissible framework derives $p$ on the shared base,
but some admissible framework refutes $p$ on that base:
\[
  \Substrate \cup \Framework' \entails \neg p .
\]
Then $p$ is not framework-variant, because no admissible framework derives $p$.
It is also not framework-invariant, because adding $p$ to the refuting base gives
\[
  \Substrate \cup \Framework' \cup \{p\} \entails \bot .
\]

The proposition-level definition tests whether a candidate proposition
can be added to the substrate while preserving compatibility with every
admissible framework.
The commitment-set definition tests whether a set of commitments is
compatible with every admissible framework.
Applied to $\SubstrateRef$,
the commitment-set definition gives the compatibility asserted by
\AssumpRef{se100.assump.ReferentialCommonGround}{Referential Common Ground}.

The neutrality constraint below requires this positive compatibility guarantee:
compatibility with every admissible framework,
including admissible frameworks not known when the substrate is designed.

\subsection*{Neutrality Conditions}


\begin{definition}[Contested Causal or Normative Proposition]
  \label{se100.def.ContestedCausalNormative}
  A causal or normative proposition $p$ is \emph{contested}
  in the relevant accountability context
  if its acceptance, rejection, interpretation, or application
  is not fixed by the referential commitments $\SubstrateRef$,
  so that it may vary across admissible frameworks.

  Let $C_{cn}$ denote the class of contested causal or normative propositions.
\end{definition}

\begin{assumption}[Contestability]
  \label{se100.assump.Contestability}
  No $p \in C_{cn}$ is guaranteed framework-invariant at design time.
\end{assumption}

Contestability does not mean that propositions in $C_{cn}$ are
false, arbitrary, unknowable, or unsupported.
It means that the substrate cannot certify their universal compatibility
across admissible frameworks at design time.
Claims that a causal or normative proposition is
well supported,
evidence-backed,
legally required,
institutionally authorized,
or supported by a cited source
must themselves be represented as attributed, provenance-bearing assertions.
They are not exceptions to the restriction on substrate-layer commitment.


\begin{assumption}[Referential Common Ground]
  \label{se100.assump.ReferentialCommonGround}
  Let $\SubstrateRef$ be the referential commitments of $\Substrate$
  (\DefRef{se100.def.ReferentialCommitments}{Referential Commitments}).
  For every admissible framework $\Framework$,
  \[
    \SubstrateRef \cup \Framework \notentails \bot .
  \]
  In particular, for every permitted attribution proposition
  $\Asserts(x,\varphi)$,
  \[
    \SubstrateRef \cup \{\Asserts(x,\varphi)\} \cup \Framework
    \notentails \bot .
  \]
  This second condition follows from the first,
  because $\SubstrateRef \entails \Asserts(x,\varphi)$ by definition;
  the argument below does not use it.
  Both conditions hold for every admissible framework,
  including admissible frameworks not known when $\Substrate$ is designed.
\end{assumption}

\AssumpRef{se100.assump.ReferentialCommonGround}{Referential Common Ground}
states that $\SubstrateRef$ is a framework-compatible commitment set and
that permitted attribution propositions whose attributional bases are fixed
by $\SubstrateRef$ remain compatible with every admissible framework.
It formalizes the scope condition on which neutrality depends:
the \emph{entire} set of adopted referential commitments must remain
compatible with each admissible framework.
Individual compatibility does not establish this requirement.

A permitted attribution proposition is entailed by the referential
commitments, so adding it supplies no independent proposition beyond
that commitment set.
This does not assert the truth of the attributed content.
Frameworks may disagree in their interpretations; the formal
failure relevant here is joint inconsistency with the adopted
referential commitments, not disagreement alone.

This is an assumption, not a general property of records.
It can fail.
\SecRef{sec:boundary} treats its failure as the condition under which
the construction does not certify neutrality.

\begin{remark}[Attribution and Common Ground]
  \label{se100.remark.AttributionCommonGround}
  A permitted attribution proposition
  $\Asserts(x,\varphi)$ commits the substrate to the attribution by $x$,
  not to the asserted proposition $\varphi$
  (\DefRef{se100.def.PermittedAttributionProposition}{Permitted Attribution Proposition}).
  An admissible framework may reject $\varphi$ without rejecting that
  $x$ asserted it.

  A permitted attribution proposition satisfies
  $\SubstrateRef \entails \Asserts(x,\varphi)$
  by definition
  (\DefRef{se100.def.PermittedAttributionProposition}{Permitted Attribution Proposition}).
  Hence adjoining it to the foundational layer adds no independent commitment
  beyond what is already determined by $\SubstrateRef$: for every admissible
  framework $\Framework$,
  \[
    \SubstrateRef \cup \{\Asserts(x,\varphi)\} \cup \Framework
    \entails \bot
    \quad\text{iff}\quad
    \SubstrateRef \cup \Framework \entails \bot .
  \]
  Therefore, by
  \AssumpRef{se100.assump.ReferentialCommonGround}{Referential Common Ground},
  \[
    \SubstrateRef \cup \{\Asserts(x,\varphi)\} \cup \Framework
    \notentails \bot .
  \]

  If an admissible framework rejects $\varphi$,
  that does not alone contest the attribution proposition $\Asserts(x,\varphi)$.
  But if admissible frameworks contest whether the attributional basis,
  for example, who asserted the claim, what was
  asserted, or what provenance identifies the assertion occurrence,
  then the attribution proposition is not permitted at the foundational layer.
  That is a boundary case, not an exception to the neutrality constraint.
\end{remark}

\begin{definition}[Interpretive Non-Commitment]
  \label{se100.def.InterpretiveNonCommitment}
  A substrate $\Substrate$ satisfies \emph{interpretive non-commitment} if it
  makes no substrate-layer commitment to any proposition that is
  framework-variant with respect to $\Substrate$ and $\Frameworks$.
  It also makes no commitment to the negation of such a proposition,
  and does not assume that negation preserves framework variance;
  explicitly, if
  $p$ is framework-variant with respect to $\Substrate$ and $\Frameworks$,
  then
  $\Substrate \notentails p$ and $\Substrate \notentails \neg p$.
\end{definition}

Interpretive non-commitment does not prevent the substrate from recording that
a framework, source, agent, institution, record, or document asserts $p$.
It prevents the substrate from asserting $p$ independently of attribution.

\begin{definition}[Extension Stability]
  \label{se100.def.ExtensionStability}
  A substrate $\Substrate$ satisfies \emph{extension stability} if, for every
  admissible framework $\Framework$,
  \[
    \Substrate \cup \Framework \notentails \bot
  \]
\end{definition}

This is pairwise compatibility between the substrate and each admissible
framework, not global consistency across all admissible frameworks.
The distinction matters: admissible frameworks may contradict one another.
A neutral substrate need not make all frameworks mutually consistent.
It must remain separately usable by each admissible framework without causing contradiction.


\begin{assumption}[Substrate Consistency]
  \label{se100.assump.SubstrateConsistency}
  Every substrate considered in this paper is internally consistent:
  $\Substrate \notentails \bot$.
\end{assumption}

This assumption excludes degenerate substrates from the domain of analysis.
An internally inconsistent substrate cannot serve as the stable shared base
presupposed by the definitions.
The explicit assumption is retained to make the baseline condition independent
of whether any admissible framework has yet entered consideration.
It prevents vacuous cases in which neutrality conditions would hold only
because there are no admissible frameworks against which the substrate is tested.


\begin{remark}[Relation Between the Two Properties]
  \label{se100.remark.PropertyRelation}
  Extension stability entails interpretive non-commitment.
  To see this, suppose that $p$ is framework-variant with respect to
  $\Substrate$ and $\Frameworks$.
  Then there are admissible frameworks $\FrameworkOne, \FrameworkTwo \in \Frameworks$
  such that
  \[
    \Substrate \cup \FrameworkOne \entails p
    \qquad\text{and}\qquad
    \Substrate \cup \FrameworkTwo \entails \neg p .
  \]

  If $\Substrate \entails p$, then
  $\Substrate \cup \FrameworkTwo \entails p$
  and
  $\Substrate \cup \FrameworkTwo \entails \neg p$,
  so
  \[
    \Substrate \cup \FrameworkTwo \entails \bot .
  \]
  If $\Substrate \entails \neg p$, then
  $\Substrate \cup \FrameworkOne \entails \neg p$
  and
  $\Substrate \cup \FrameworkOne \entails p$,
  so
  \[
    \Substrate \cup \FrameworkOne \entails \bot .
  \]
  In either case, extension stability fails.
  Therefore, if extension stability holds, the substrate cannot commit to
  either side of a framework-variant proposition.

  This paper uses only this direction.
  The two properties are retained under separate names because they identify
  distinct design failures:
  substrate-layer commitment on disputed content and breakdown of layerability.
\end{remark}

\section{The Neutrality Constraint}
\label{sec:neutrality}

\begin{definition}[Neutrality by Design]
  \label{se100.def.NeutralityByDesign}
  A substrate $\Substrate$ is \emph{neutral}
  when it satisfies interpretive non-commitment
  (\DefRef{se100.def.InterpretiveNonCommitment}{Interpretive Non-Commitment})
  and extension stability
  (\DefRef{se100.def.ExtensionStability}{Extension Stability}).

  It is \emph{neutral by design} when some permitted
  design-time basis establishes that neutrality for the full
  intrinsic class of admissible frameworks, including frameworks
  not known when the substrate is constructed.
  Truth of the neutrality condition and possession of such a
  permitted basis are distinct claims.
\end{definition}

The distinction is between a semantic property and a design-time guarantee.
Neutrality is the semantic property that interpretive non-commitment and
extension stability hold relative to $\Frameworks$.
Neutrality by design is stronger:
it requires a permitted design-time basis that establishes semantic neutrality.
The abstract guarantee-system interface does not specify which bases
are permitted or independently enforce a prohibition on framework enumeration.
A concrete realization must justify its permitted bases and
the guarantees they establish.

\begin{constraint}[Neutrality]
  \label{se100.constraint.Neutrality}
  Let $\Substrate$ be a substrate with referential commitments
  $\SubstrateRef$.
  Assume
  \AssumpRef{se100.assump.ReferentialCommonGround}{Referential Common Ground}.
  If every substrate-layer commitment is either a member of
  $\SubstrateRef$ or a permitted attribution proposition
  (\DefRef{se100.def.PermittedAttributionProposition}
  {Permitted Attribution Proposition}),
  then $\Substrate$ satisfies extension stability and
  interpretive non-commitment for every admissible framework.
  Thus it is semantically neutral.

  This result does not, without a further permitted
  design-time-guarantee basis, establish neutrality by design.
  Nor does it assert that every neutral substrate must satisfy
  this specific restriction.
\end{constraint}

The result establishes conditional \emph{sufficiency}, not an
unconditional biconditional or an exhaustive characterization of all
possible neutral architectures.

\begin{remark}[Characterization as an Explicit Hypothesis]
  \label{se100.remark.CharacterizationHypothesis}
  The biconditional (under Contestability and Referential Common
  Ground, neutrality by design holds if and only if the foundational
  layer is restricted to the permitted classes) is not derived here.
  Its forward direction is the by-design form of
  \ConstRef{se100.constraint.Neutrality}{Neutrality}, which requires a
  permitted design-time basis; its reverse direction is the converse
  discussed below under limits of necessity.
  The accompanying Lean formalization represents the biconditional,
  together with the exclusion of object-level causal or normative
  commitments by that restriction, as an explicit hypothesis on a
  realization, named \texttt{CharacterizationHypothesis}.
  It is an assumption about the realization, not a result of this paper.
\end{remark}

\emph{Justification.}
By the foundational-layer restriction, every substrate commitment is
either a referential commitment or a permitted attribution.
By definition, each permitted attribution is entailed by the
referential commitments.
Consequently, the complete substrate commitment theory adds no
independently asserted proposition beyond those commitments.
Under Referential Common Ground, the referential commitment theory
remains jointly consistent when combined with each admissible framework.
By the structural consequence laws (including generalized cut),
the complete substrate commitment theory therefore also remains
consistent with each admissible framework.
This proves extension stability.
Extension stability entails interpretive non-commitment by
\RemRef{se100.remark.PropertyRelation}{Relation Between the Two Properties}.
Hence semantic neutrality follows.

The formalization also contains a concrete model showing that these notions
are not vacuous:
two admissible frameworks that disagree about a contested proposition,
a substrate that abstains from it and is neutral, and
a substrate that commits to it and is not neutral,
because it becomes inconsistent with one of the frameworks.

The argument does not provide a semantic or empirical warrant for
the referential commitments or the attributed assertion records.
Those are premises supplied by the intended realization.
It also does not construct a permitted basis that establishes neutrality
at design time.

The Lean formalization proves semantic neutrality from the
foundational-layer restriction and Referential Common Ground alone.
It does not establish neutrality by design, which additionally
requires a permitted design-time basis that establishes neutrality
in the specified design-time guarantee system.

\emph{Limits of necessity.}
The converse is not derived from Contestability and Referential Common Ground alone.
It is imposed as a hypothesis on a realization
(\texttt{CharacterizationHypothesis} in the formalization).
The fact that a proposition is outside the two designated classes
does not establish its incompatibility with any admissible framework.
In particular,
\AssumpRef{se100.assump.Contestability}{Contestability}
does not supply a refuting framework for every proposition outside
those classes.
Any stronger necessity result would require explicit additional
hypotheses or a narrower definition of the intended design discipline.
Defining admissible design bases by the two permitted classes is a
stipulated design restriction, not an independent impossibility proof.

Likewise, the layer restriction does not by itself show that no
object-level causal or normative proposition is among
$\SubstrateRef$.
That separation depends on independently justified referential-fixing
semantics and interpretation of the propositions involved.

\section{Mechanism: Reification of Interpretive Claims}
\label{sec:mechanism}

The mechanism used by the permitted-layer construction is reification.
Consider a substrate that adopts an object-level interpretive proposition
incompatible with some admissible framework:
\[
  \Substrate \entails \varphi
\]
Under a fixed and warranted attributional basis,
the permitted-layer construction instead represents
that some framework, source, agent, institution, record, or document $x$
asserts the proposition:
\[
  \Substrate \entails \Asserts(x,\varphi)
\]
The substrate commits to the attribution proposition
$\Asserts(x,\varphi)$,
not to the asserted proposition $\varphi$
(\DefRef{se100.def.SubstrateCommitment}{Substrate-Layer Commitment}).

This allows the record to preserve accountability-relevant content without
making the shared layer partisan among admissible frameworks.
The substrate may record:
the decision that was made;
the person or entity affected;
the time of the decision;
the instrument or system involved;
the policy or rule cited;
the source that asserted the claim;
the text of the claim;
the causal or normative proposition asserted;
and the provenance of the assertion.
The substrate must not convert the asserted proposition into a
substrate-layer commitment to that proposition.

\section{Worked Example}
\label{sec:example}

\begin{example}[Reification Fragment]
  \label{se100.example.ReificationFragment}
  Consider a decision record concerning an automated eligibility denial.
  The record involves a subject $u$, a decision $d$, a model or instrument $m$,
  a timestamp $t$, an institutional policy $r$, and
  a claim asserted by a source $x$ (for example, a framework).
\end{example}

The example is conditional: it presupposes a referential realization
that justifies the subject, decision, instrument, and policy as the
intended referents, fixes the source $x$ and its assertion occurrences
as the attributional basis for the attribution propositions below,
and establishes compatibility of the resulting
commitment set with every admissible framework.
The symbols and identifiers do not by themselves establish
individuality, denotation, co-reference, or persistence.
The example illustrates layer separation under those assumptions;
it is not independent evidence that they hold in a deployed system.

\textbf{Non-neutral form.}
The substrate makes object-level causal or normative commitments
(\DefRef{se100.def.ObjectLevelCausalNormativeCommitment}
{Object-Level Causal or Normative Commitment}):
\[
  \Substrate \entails \pred{caused\_by}(\pred{score}(u,m), \pred{denial}(d))
  \qquad
  \Substrate \entails \pred{justified\_by}(\pred{denial}(d), \pred{policy}(r))
\]
If the resulting commitments are inconsistent with an admissible
framework, extension stability fails and the substrate is not neutral.
Disagreement or unresolved evidentiary warrant by itself does not prove
formal inconsistency; it prevents this example from being certified
without the required compatibility analysis.

\textbf{Neutral form.}
The substrate makes referential commitments:
\begin{align*}
  \Substrate & \entails \pred{Subject}(u)
             & \Substrate                    & \entails \pred{occurred\_at}(d,t)     \\
  \Substrate & \entails \pred{Decision}(d)
             & \Substrate                    & \entails \pred{involved}(d,u)         \\
  \Substrate & \entails \pred{Instrument}(m)
             & \Substrate                    & \entails \pred{used\_instrument}(d,m) \\
  \Substrate & \entails \pred{Policy}(r)
             & \Substrate                    & \entails \pred{cites}(d,r)
\end{align*}
and records the contested content as attribution propositions:
\[
  \Substrate \entails \Asserts(x,\;
  \pred{caused\_by}(\pred{score}(u,m), \pred{denial}(d)))
\]
\[
  \Substrate \entails \Asserts(x,\;
  \pred{justified\_by}(\pred{denial}(d), \pred{policy}(r)))
\]
The substrate preserves that source $x$
made these causal and normative claims.
It commits to the attribution propositions,
not to the asserted propositions.

\textbf{Predicate-level reading.}
The neutral record contains the
decision, subject, timestamp, identifiers, institutional artifacts, and
provenance-bearing attribution propositions.
It contains no unattributed top-level content whose meaning is equivalent to
the non-neutral commitments above.
This holds regardless of field naming:
a field named \texttt{reason}, \texttt{cause},
\texttt{basis}, \texttt{risk}, \texttt{fault}, or \texttt{violation}
is not prohibited by name.
The question is whether, in the schema and use context,
the field functions as a substrate-layer commitment to an object-level causal or normative proposition.
For example, a cited policy identifier or recorded model score may be referential content,
while an unattributed assertion that the policy justified the decision,
the score caused the denial, or the subject violated a rule is
causal or normative content on the substrate's own account.
Causal or normative vocabulary may appear;
the test is whether the substrate makes a substrate-layer commitment
to the asserted causal or normative proposition itself,
rather than to an attribution proposition about that proposition
(\NoteRef{se100.note.CausalNormative}{Causal and Normative Content as Primitive
  Classifications}).

The neutral form supports multiple downstream interpretations without
requiring the substrate to choose among them.
An institutional framework may treat the attributed claim as sufficient
for its purpose.
An audit framework may question whether the cited policy applied.
A claimant-side framework may contest the causal proposition.
A regulatory framework may evaluate whether the attributed claim satisfies
a disclosure rule.
The substrate remains usable by all of them
because it preserves the shared referential base and the attribution
propositions without making an object-level substrate-layer commitment
to the contested interpretation.

\section{Boundary Condition: When Neutrality Is Not Certified}
\label{sec:boundary}

Neutrality in the permitted-layer construction depends on
\AssumpRef{se100.assump.ReferentialCommonGround}{Referential Common Ground}.
This assumption concerns \emph{joint compatibility} of the complete
referential commitment set with each admissible framework.
It can fail when an adopted referential criterion, a claimed co-reference
relation, a persistence judgment, or an attributional basis conflicts
with commitments of an admissible framework.

Disagreement about the interpretation of an event does not by itself
show that Referential Common Ground fails.
Nor does the absence of a positive co-reference warrant establish
that two references are to distinct referents.
The relevant test is whether the adopted foundational propositions
are warranted as fixed and remain jointly compatible with every
admissible framework.

For example, two independently produced records may be attributed
to the same institutional event or to different events.
A shared identifier, timestamp, source attribution, or agreement
on selected fields may be insufficient to decide the question.
If the substrate commits to their co-reference without an independently
justified denotation and identity criterion, it has not established the
premises required for the neutrality result.
If the adopted co-reference proposition conflicts with an admissible
framework, the common-ground condition fails for that realization.

Similarly, a framework may reject an asserted proposition $\varphi$
while accepting that a source $x$ asserted $\varphi$.
That need not defeat compatibility.
But the attribution itself cannot be treated as permitted merely
because a provenance field exists:
its source, assertion occurrence, content, and record basis
must be semantically fixed by the referential commitments.

When these premises cannot be established, the present construction
does not certify neutrality at the proposed foundational layer.
It does not follow that accountability is impossible or that no
alternative shared structure could be neutral.
A system may instead represent contested referential judgments
as attributed claims while withholding their adoption as common
substrate commitments.
Such a revised construction requires its own compatibility analysis.

\section{Scope and Limits}
\label{sec:non-goals}

\textbf{This paper does not propose a data format.}
A neutral substrate may be implemented using relational databases,
graph structures, event logs, RDF-like representations,
document stores, schemas, or application-specific formats.
The neutrality constraint applies to the substrate-layer commitments
made by the foundational layer,
not to any serialization or storage technology.

\textbf{This paper does not adjudicate among frameworks.}
Admissibility identifies minimum conditions for participation in the shared
record:
internal consistency,
evidentiary grounding,
and a documented interpretive function.
It does not decide which admissible framework should prevail in a dispute.

\textbf{This paper does not claim a neutral admissibility gate.}
As stated in \SecRef{sec:definitions},
the admissibility gate embeds design choices about what counts as
a framework that can enter the shared record.
The neutrality defined here is conditional on those choices.
The gate's role is to make participating frameworks identifiable,
scoped, and citable,
not to determine which framework is correct.

\textbf{This paper does not solve accountability generally.}
Neutrality is a structural condition for
shared records under persistent disagreement.
It does not determine liability, responsibility, institutional authority,
causal truth, legal compliance, policy adequacy, or
which interpretation should govern.
It makes such disputes representable without
forcing the shared substrate to decide them.

\textbf{This paper does not prohibit causal or normative content.}
The permitted-layer construction records contested object-level
interpretive claims as attributed assertions rather than adopting them
as unqualified foundational commitments.
The formal sufficiency theorem does not independently establish that
all object-level causal or normative propositions are excluded from
the referential commitment set.
Causal and normative claims remain representable by attribution with provenance.

\textbf{This paper does not establish referent-specific identity criteria.}
The referential regime is an abstract interface for individuation,
co-reference, and persistence.
It does not justify which observations constitute one whole,
when two records denote the same referent,
or what licenses strict identity across settings.
Such judgments require independent grounds, interpretation conditions,
and examination of relevant counterexamples.

\textbf{This paper does not provide a general compatibility algorithm.}
Compatibility concerns the complete commitment set and every admissible
framework.
The theory imposes no finite-enumeration or decision procedure,
nor does it assume that failure to entail a proposition warrants
its negation.

\textbf{This paper does not derive neutrality by design from neutrality.}
Semantic neutrality can follow from the foundational restriction under
Referential Common Ground.
A design-time claim requires an independently
supplied permitted guarantee basis.
No such concrete basis is established for the worked example.

\section{Discussion}
\label{sec:discussion}

The constraint is most useful in applied ontology settings
where reference is shared but interpretation remains persistently contested.
Examples include
litigation,
forensic discovery,
audit,
regulatory review,
institutional accountability records, and
multi-party AI-governance records.

The permitted-layer discipline separates justified referential
commitments from attributed interpretive assertions.
It requires independent grounding of the referential base
and a permitted design-time basis for any claim of neutrality by design.
This discipline matters in systems that combine records from multiple
institutions, jurisdictions, tools, or governance contexts,
where causal and normative terms often appear as ordinary metadata.
Neutral design does not prohibit such terms.
It asks what the substrate is committing to
(\DefRef{se100.def.SubstrateCommitment}{Substrate-Layer Commitment}).
If the substrate asserts that a denial was caused by a score,
that a policy justified a decision,
or that a party violated a norm,
the substrate has made an object-level causal or normative commitment.
If it records that a specified source made that claim under documented
provenance,
the substrate has preserved the claim without adopting it.

The constraint has practical value only where referential common ground holds.
In these settings, parties share enough reference
to identify which record,
which decision, which subject, and which instrument are at issue while disputing,
often persistently, what caused the outcome and whether it was justified.
This may be a common condition of records under dispute.
Hard accountability cases, including disputed event identity
and institutional persistence, may lack warranted referential common ground.
Those are the boundary cases of \SecRef{sec:boundary}.
The paper provides a conditional sufficient construction where the
joint-compatibility and fixing premises hold, not a universal criterion
for determining when they hold or a proof that alternatives are impossible.

\section{Conclusion}
\label{sec:conclusion}

A shared accountability record must often serve parties who disagree
persistently about causation, norms, responsibility, legitimacy, or
institutional interpretation.
If its foundational commitments are incompatible with an admissible
framework, the record fails extension stability and therefore neutrality.
A merely contested interpretation has not thereby been proved incompatible;
the permitted-layer construction withholds unqualified adoption until
the required referential and compatibility premises are justified.

This does not make the record empty.
It makes the record usable for accountability.
The substrate can preserve who said what,
about which entity, event, decision, instrument, policy, or artifact,
under what provenance,
and within what scope.
It must not convert an attributed interpretation into
an object-level causal or normative commitment of the substrate.

Further work must justify referential commitments for intended
referent kinds, establish their joint compatibility, and specify
permitted design-time guarantee bases and application-level checks.

The established result is conditional: under Referential Common Ground,
restricting foundational commitments to the permitted classes yields
semantic neutrality.
When that assumption cannot be warranted,
the construction provides no neutrality certificate for the proposed
referential basis.
It does not resolve contested reference merely
by recording identifiers or provenance, and it does not claim to
exclude every other neutral architecture.

\section*{Statements and Declarations}

\subsection*{Acknowledgments} The author thanks the anonymous reviewers
of an earlier version of this work for comments that helped improve
the framing, organization, and presentation of the work.

\subsection*{Author Contributions}
The author is the sole contributor to this work and
is responsible for all aspects of the research, authorship, and publication.

\subsection*{Use of AI-Assisted Tools}
AI-assisted tools were used for editing, formatting, and consistency checking.
The author reviewed all suggestions and is solely responsible for the content.

\subsection*{Declaration of Conflicting Interest}
The author declares no potential conflicts of interest with respect to the
research, authorship, and publication of this work.

\end{document}